\documentclass[final]{clv2025}
\usepackage{algorithm}
\usepackage{algorithmic}

\jvol{vv}
\jnum{nn}
\jyear{2025}

\dochead{Short Paper} 

\pageonefooter{Action editor: \{action editor name\}. Submission received: DD Month YYYY; revised version received: DD Month YYYY; accepted for publication: DD Month YYYY.}

\usepackage{amsmath}
\usepackage{booktabs}
\usepackage{amsfonts}
\usepackage{amssymb}
\usepackage{dsfont}  
\runningtitle{DRE}
\runningauthor{Kun et.al}

\begin{document}

\title{DRE: An Effective Dual-Refined Method for Integrating Small and Large Language Models in Open-Domain Dialogue Evaluation}

\author{Kun Zhao$^{1}$, Bohao Yang$^{2}$, Chen Tang$^{2}$, Siyuan Dai$^{1}$, Haoteng Tang$^{4}$, Chenghua Lin$^{*,2}$, Liang Zhan\thanks{Corresponding authors}$^{1}$}

\affilblock{
    \affil{Department of Electrical and Computer Engineering, University of Pittsburgh, US\\\quad \email{{kun.zhao, liang.zhan}@pitt.edu}}
    \affil{Department of Computer Science, The University of Manchester, UK\\\quad
    \email{chenghua.lin@manchester.ac.uk}}
    \affil{Department of Computer Science, University of Texas Rio Grande Valley, US}
}

\maketitle

\begin{abstract}
Large Language Models (LLMs) excel at many tasks but struggle with ambiguous scenarios where multiple valid responses exist, often yielding unreliable results. Conversely, Small Language Models (SLMs) demonstrate robustness in such scenarios but are susceptible to misleading or adversarial inputs. We observed that LLMs handle negative examples effectively, while SLMs excel with positive examples. To leverage their complementary strengths, we introduce SLIDE (Small and Large Integrated for Dialogue Evaluation), a method integrating SLMs and LLMs via adaptive weighting. Building on SLIDE, we further propose a Dual-Refinement Evaluation (DRE) method to enhance SLM-LLM integration: (1) SLM-generated insights guide the LLM to produce initial evaluations; (2) SLM-derived adjustments refine the LLM’s scores for improved accuracy. Experiments demonstrate that DRE outperforms existing methods, showing stronger alignment with human judgment across diverse benchmarks. This work illustrates how combining small and large models can yield more reliable evaluation tools, particularly for open-ended tasks such as dialogue evaluation. 
\end{abstract}

\section{Introduction}
Open-domain dialogue generation represents a significant research focus within Natural Language Generation (NLG)~\cite{zhao2023evaluating, xiao2023evaluating, ye2024uncertainty, tang2023enhancing, loakman2023iron, yang2024improving, dai2024constrained}. Evaluating such dialogues is challenging due to the one-to-many nature of the problem, wherein a single conversational context admits multiple plausible, semantically appropriate responses. Adversarial negative responses exhibiting lexical overlap with contexts further complicate assessment~\cite{Sai2020ImprovingDE}.

Existing evaluation methods rely on word-overlap metrics (e.g., BLEU \cite{papineni-etal-2002-bleu}, ROUGE \cite{lin-2004-rouge}, METEOR \cite{banerjee-lavie-2005-meteor}) or semantic-embedding approaches (e.g., BERTScore \cite{Zhang2020BERTScoreET}, BARTScore \cite{yuan2021bartscore}), which compute similarity between generated responses and context/reference sets. These methods inadequately address the one-to-many problem. While large language models (LLMs) pre-trained on extensive datasets demonstrate strong performance in evaluation tasks~\cite{fu-etal-2024-gptscore, liu-etal-2023-g, Kocmi2023LargeLM, Chiang2023CanLL}, their broad pre-training scope limits task-specific knowledge. Consequently, they may exhibit hallucinations during task execution, leading to suboptimal evaluation performance~\cite{wang-etal-2024-improving-text, Lin2023LLMEvalUM, wang-etal-2024-large-language-models-fair}. For instance, \citet{Lin2023LLMEvalUM} proposed LLM-EVAL for open-domain dialogue evaluation using multiple LLMs, identifying significant prompt sensitivity that artificially constrains metric effectiveness and causes evaluation quality to fluctuate with prompt formulation, resulting in unreliable and non-robust scoring. Additionally, \cite{wang-etal-2024-large-language-models-fair} state that without additional support, LLMs alone do not constitute equitable evaluators.

In addition, LLMs encounter significant limitations in producing reliable evaluations, particularly when distinguishing between high-quality positive responses and adversarial negative ones. While LLMs exhibit moderate accuracy in classifying adversarial responses, their ability to differentiate among multiple plausible positive responses—a core challenge in one-to-many open-domain dialogue systems—remains limited. In such scenarios, semantically distinct yet contextually appropriate responses often diverge markedly from reference answers, complicating LLM-based evaluation. Small language models (SLMs), though cost-effective, exhibit inherent biases toward positive responses due to their narrower training scope. However, deploying SLMs is constrained by their dependency on large-scale, topic-diverse training data. This requirement exacerbates challenges in open-domain dialogue applications, where data scarcity and insufficient topic coverage hinder both implementation efficacy and scalability.

To address the issues outlined above, we propose a hybrid approach integrating SLMs and LLMs to resolve the one-to-many challenge and mitigate data scarcity via LLM-based augmentation. We investigate methods leveraging prior knowledge and domain-specific constraints to synergize the complementary strengths of both models. This approach, named SLIDE (Small and Large Integrated for Dialogue Evaluation), optimizes the balance between SLMs and LLMs for dialogue evaluation.
First, an SLM is trained using contrastive learning to minimize the cosine distance between positive responses and context embeddings while maximizing the distance for adversarial negative embeddings. After contrastive fine-tuning, the SLM effectively distinguishes between response types. Response representations are then categorized as robust or non-robust vectors, with only robust vectors retained. The context-response distance and classification probability are subsequently computed. These values are combined to form the SLM evaluation score, which is integrated with the LLM's output to produce the final evaluation metric.

Building on SLIDE, we propose the Dual-Refinement Evaluation method (DRE), a novel framework that integrates SLMs and LLMs through a self-learned refinement loop. Following the same SLM training procedure as SLIDE, DRE implements two distinct constraint mechanisms:(1) \textbf{Interior refinement:} SLM outputs are integrated into prompts to guide LLM evaluations, producing both a preliminary score and an influence score quantifying the SLM's impact; (2) \textbf{Exterior refinement:} The LLM score is refined using a coefficient derived from context-response distance, classification probability, and the influence score.
These stages collectively generate the final evaluation metric. Experiments across diverse LLMs confirm DRE's robustness, demonstrating significant performance improvements in dialogue evaluation.

Our contributions can be summarised as the following:
\begin{itemize}
    \item We explore and analyze various methods, including firm-constraint-based approaches (SLIDE Algorithm) and self-learned constraint mechanisms (DRE Algorithm), to transfer task-specific knowledge from SLMs to mitigate hallucination issues and thereby improve LLM performance.  To the best of our knowledge, this is the first approach aimed at enhancing the performance of LLMs in evaluating open-domain dialogues through this dual-refinement process.
    \item We design a novel classifier that categorizes dialogue responses as either positive or negative, by integrating embedding distances and probability values, achieving state-of-the-art performance on multi-reference datasets.
    \item We augment existing dialogue evaluation datasets by introducing multiple positive and adversarial negative responses. This enhanced dataset can be leveraged for fine-tuning open-domain SLM models.
    \item We conduct extensive experiments to validate the effectiveness of the proposed evaluation framework, demonstrating that DRE achieves state-of-the-art performance on open-domain dialogue evaluation benchmarks. Furthermore, our investigation into the integration of SLMs and LLMs offers valuable insights for future research on combining these models to enhance dialogue evaluation.
\end{itemize}

\section{Related Work}
\subsection{Dialogue Evaluation Metrics}

Traditional $n$-gram-based evaluation metrics such as BLEU~\cite{papineni-etal-2002-bleu}, ROUGE~\cite{lin-2004-rouge}, and METEOR~\cite{banerjee-lavie-2005-meteor} assess the overlap of words between candidate responses and a reference. Conversely, embedding-based metrics like Extrema~\cite{forgues2014bootstrapping} and BERTScore~\cite{Zhang2020BERTScoreET} transform the responses and references into high-dimensional representations to evaluate semantic similarities. Recent years have witnessed an increasing interest in developing trainable metrics. For instance, RUBER~\cite{Tao2018} evaluates the similarity between the generated response, its context, and the actual reference. DEB~\cite{Sai2020ImprovingDE}, a BERT-based model pre-trained on extensive Reddit conversation datasets, was introduced to enhance evaluation effectiveness. The Mask-and-Fill approach~\cite{Gupta2021SynthesizingAN}, utilizing Speaker Aware (SA)-BERT~\cite{Gu2020SpeakerAwareBF}, aims to better understand dialogues. MDD-Eval~\cite{Zhang2021MDDEvalSO} was developed to evaluate dialogue systems across various domains by using a teacher model to annotate dialogues from different domains, thus supporting the training of a domain-agnostic evaluator. This method, however, relies on human labels and additional training data, which our proposed approach does not require. CMN~\cite{zhao2023evaluating} was designed to effectively address the one-to-many nature of open-domain dialogue evaluation in the latent space, yet it does not account for adversarial negative examples. In this paper, we design two novel methods to evaluate multiple responses, which contain adversarial negative responses and positive responses.

\subsection{LLM-based Evaluators}

The utilization of LLMs for evaluation purposes has garnered significant interest due to their robust capabilities across a variety of tasks. GPTScore, introduced by \cite{fu-etal-2024-gptscore}, offers a multi-faceted, tailor-made, and training-independent method for evaluation. Preliminary investigations into the effectiveness of evaluators based on LLMs have been conducted by \cite{wang-etal-2023-chatgpt}. Additionally, \cite{Kocmi2023LargeLM} have employed GPT models to assess the quality of machine translation. Moreover, \cite{liu-etal-2023-g} developed GPT-EVAL, which leverages GPT-4 to evaluate performance across multiple tasks, including dialogue response generation, text summarization, data-to-text generation, and machine translation. Despite these advancements, the application of LLM-based metrics for assessing adversarial negative responses within open-domain dialogue scenarios remains unexplored. Additionally, because of the hallucination issues that exist, they also have many limitations. In order to let the LLM-based metrics become more robust, we utilized the SLM to refine the LLM and thus alleviate hallucination issue of LLMs to make it become a more robust evaluator.

\section{Methodology}
\subsection{Preliminaries}
\subsubsection{Transformers}
For a sentence $x_i$, assuming it contains j tokens, the embedding process is defined as follows:

\begin{align} 
    \mathbf{X_i} = [\mathbf{E}(x_{i,1}); \mathbf{E}(x_{i,2}); \dots; \mathbf{E}(x_{i,j})] \in \mathbb{R}^{j \times d}
\end{align}

where $\mathbf{X_i}$ is the embedded representation of the input sequence, $j$ is the sequence length, and $d$ is the embedding dimension. In this paper, $d$=768.

Since the Transformer inherently lacks sequential information, , it is necessary to incorporate positional encodings into the input embeddings. Typically, this is achieved by applying sine and cosine functions to represent positional information:

\begin{align} 
    \mathbf{PE}_{(pos, 2j)} &= \sin \left( \frac{pos}{10000^{2j/d}} \right) \nonumber\\ 
    \mathbf{PE}_{(pos, 2j+1)} &= \cos \left( \frac{pos}{10000^{2j/d}} \right) 
\end{align}

The final input representation is then:

\begin{align} 
    \mathbf{Z_{i,0}} &= \mathbf{X_i} + \mathbf{PE}
\end{align}

Subsequently, the final input representation is fed into the Multi-Head Attention layers, which form the core component of the Transformer encoder. At each layer, the input matrix $\mathbf{Z_{i,l-1}}$ is processed through multiple attention heads. Each attention head computes the Query, Key, and Value vectors for the attention mechanism:

\begin{align} 
    \mathbf{Q}_k &= \mathbf{Z}_{i,l-1} \mathbf{W}_k^Q, \nonumber\\ 
    \mathbf{K}_k &= \mathbf{Z}_{i,l-1} \mathbf{W}_k^K, \nonumber\\ 
    \mathbf{V}_k &= \mathbf{Z}_{i,l-1} \mathbf{W}_k^V
\end{align}
where $\mathbf{Q}_k, \mathbf{K}_k, \mathbf{V}_k$  are learned weight matrices.

The self-attention scores are computed as:

\begin{align} 
    \text{Attention}(\mathbf{Q}, \mathbf{K}, \mathbf{V}) = \text{softmax}\left( \frac{\mathbf{Q} \mathbf{K}^T}{\sqrt{d_k}} \right) \mathbf{V}
\end{align}

The multi-head attention mechanism concatenates the outputs of the attention heads and applies a linear transformation:
\begin{align} 
    \text{MultiHead}(\mathbf{Q}, \mathbf{K}, \mathbf{V}) = \text{Concat}(\text{head}_1, \dots, \text{head}_h) \mathbf{W}^O
\end{align}
where $h$ is the number of heads, and $\mathbf{W}^O$is the linear transformation matrix.

The output of the attention layer is passed through a feed-forward neural network consisting of two linear layers with a non-linear activation function (such as ReLU):

\begin{align} 
    \text{FFN}(\mathbf{z}) = \max(0, \mathbf{z} \mathbf{W}_1 + b_1) \mathbf{W}_2 + b_2
\end{align}

After each sub-layer, residual connections and layer normalization are applied:

\begin{align} 
    \mathbf{Z}_l' &= \text{LayerNorm}(\mathbf{Z}_{l-1} + \text{MultiHead}(\mathbf{Q}, \mathbf{K}, \mathbf{V}))
    \nonumber\\ 
    \mathbf{Z}_l &= \text{LayerNorm}(\mathbf{Z}_l' + \text{FFN}(\mathbf{Z}_l'))
\end{align}
where $\mathbf{Z}_l$ is the output of the $l-th$ layer. Typically, there are many sub-layers and we will select the output of the last sub-layer as the representation vector of the input sentence. Through these process, we will acquire the hidden state of $<x_i^c, x_i^p, x_i^a>$, noting as follows:
\begin{align} 
    h_i^c &= \mathrm{Encoder}(x_i^c) \nonumber\\ 
    h_i^p &= \mathrm{Encoder}(x_i^p) \nonumber\\ 
    h_i^a &= \mathrm{Encoder}(x_i^a) 
\end{align}

\subsubsection{Constrastive Learning}

Contrastive representation learning aims to learn an embedding space where similar sample pairs remain proximate while dissimilar pairs are distantly separated. This approach applies to both supervised and unsupervised settings, serving as a powerful self-supervised method when labels are unavailable.

Early contrastive loss functions utilized only one positive and one negative sample per input. Contemporary objectives, however, incorporate multiple positive and negative pairs within batches. Contrastive loss represents one of the earliest training objectives for deep metric learning in this paradigm.

Given input samples ${ \mathbf{x}_i }$ with corresponding labels $y_i \in {1, \ldots, L}$ across $L$ classes, we seek an embedding function $f\theta: \mathcal{X} \rightarrow \mathbb{R}^d$ that maps samples such that intra-class embeddings exhibit high similarity while inter-class embeddings demonstrate maximal separation. Consequently, the contrastive loss operates on pairs $(\mathbf{x}_i, \mathbf{x}_j)$, minimizing embedding distance for same-class pairs and maximizing it otherwise.
\begin{align}
\mathcal{L}_{\text{cont}}(\mathbf{x}_i, \mathbf{x}_j, \theta) = 
\mathds{1}_{[y_i = y_j]} \left\| f_\theta(\mathbf{x}_i) - f_\theta(\mathbf{x}_j) \right\|_2^2 + 
\mathds{1}_{[y_i \neq y_j]} \max \left( 0, \epsilon - \left\| f_\theta(\mathbf{x}_i) - f_\theta(\mathbf{x}_j) \right\|_2 \right)^2
\end{align}

where $\epsilon$ is a hyperparameter defining the minimum inter-class separation margin.

Triplet loss \cite{schroff2015facenet}, introduced in FaceNet for face verification, optimizes embeddings by simultaneously reducing anchor-positive distance and increasing anchor-negative distance. Given an anchor $\mathbf{x}$, a positive sample $\mathbf{x}^+$ (same class), and a negative sample $\mathbf{x}^-$ (different class), the loss function enforces:

\begin{align}
    \mathcal{L}_{\text{triplet}}(\mathbf{x}, \mathbf{x}^+, \mathbf{x}^-) = 
\sum_{\mathbf{x} \in \mathcal{X}} \max \left( 0, \| f(\mathbf{x}) - f(\mathbf{x}^+) \|_2^2 - \| f(\mathbf{x}) - f(\mathbf{x}^-) \|_2^2 + \epsilon \right)
\end{align}

where the margin hyperparameter $\epsilon$ enforces a minimum separation between positive and negative pair distances.

\subsection{Model Architecture}

\begin{figure*}[ht]
\small
\centering 
\includegraphics[scale=0.4]{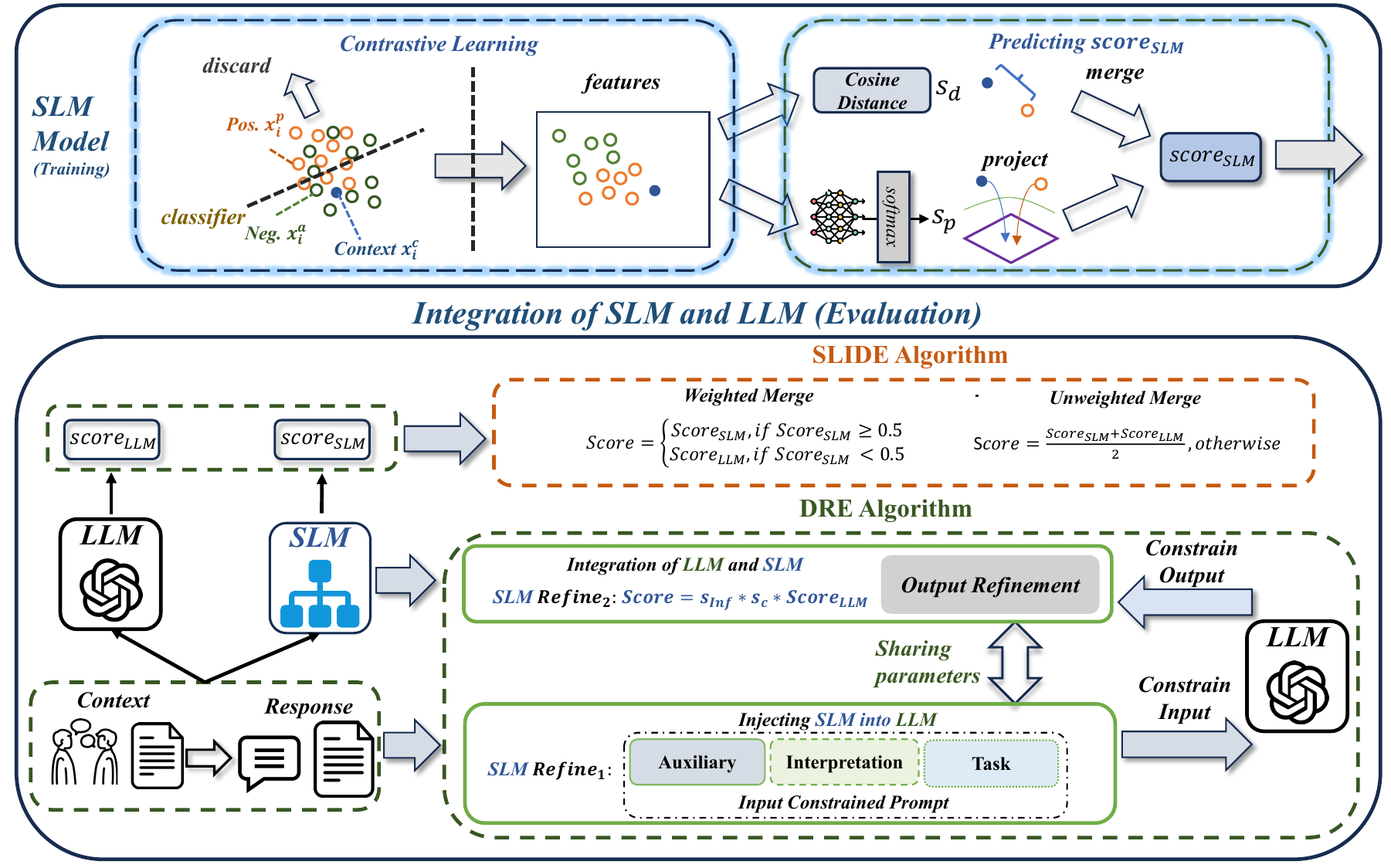}
\caption{The architecture of the proposed models involves first utilizing an SLM, trained through contrastive learning, to compute the cosine distance between the context and the responses, as well as the probability of a response being positive. We then propose two algorithms to leverage this information. The first, SLIDE, directly integrates SLM outputs with the LLM score. The second, DRE, incorporates SLM information into the LLM prompt to compute a new coefficient, $s_c$, and an influence score $s_{Inf}$, refining the LLM score. The LLM then generates $score_{LLM}$, along with an influence score $S_{Inf}$, indicating the impact of the SLM on the LLM. The final score is derived by refining the LLM score through two stages of SLM integration.}
\label{fig:DRE}
\end{figure*}

As shown in Figure~\ref{fig:DRE}, we propose two methods for evaluating open-domain dialogues. During the training phase, the focus is on training SLMs to accurately classify positive and adversarial negative responses. In the evaluation phase, we apply different methods to integrate LLMs and SLMs, thereby completing the evaluation process.

\subsection{Training Process}

To effectively distinguish between negative and positive responses, we design a contrastive learning loss function that increases the semantic distance between the context and negative responses while reducing the distance between the context and positive responses. To achieve this goal, we employ Sentence-Transformer~\cite{reimers2019sentence} to encode the context and responses. Specifically, the model is trained to bring the embeddings of positive responses closer to the context embeddings, while pushing the embeddings of adversarial negative responses further away. Given a triplet  $<x_i^c, x_i^p, x_i^a>, $ where  $1 \leq i \leq n$, $x_i^c$ denoting context, $x_i^p$ is positive responses, and $x_i^a$ represents adversarial negative responses. Each sentence is first transformed into a d-dimensional matrix through the embedding process. The objecitive of the training process is to minimize the distance between $x_i^p$ and $x_i^c$ while maximizing the separation between $x_i^a$ and $x_i^c$. The loss function is defined as follows:
\begin{align} 
    \mathcal{L} = \mathrm{max}(||h_i^c - h_i^p||   
    - ||h_i^c - h_i^a|| + \mathrm{margin}, 0)
\end{align}
in which $\mathrm{margin}$ is a hyperparameter, which is set to 0.5. $h_i$ is the hidden state of the $h_i$, whilst $||h_i^c - x_i^p|| - ||h_i^c - x_i^a||$ refers to the cosine distance.

To enhance the precision of classification, we undertake a disentanglement process. The response embeddings are disentangled into two different sub-representations: robust embeddings and non-robust embeddings. 

The robust embedding can be interpreted as a salient feature for classification, whilst the non-robust embedding constitutes noise that could act as an interfering element, potentially leading to wrong predictions.

We denote the robust and non-robust embeddings of positive responses $h_i^p$ as $\{h_i^{pr}, h_i^{pn}\}$, respectively. On the other hand, $\{h_i^{ar}, h_i^{an}\}$ represents the robust and non-robust embeddings of adversarial negative responses $h_i^a$. It is imperative that the robust and non-robust embeddings within both  positive responses and adversarial negative responses maintain clear distinctions.
Similarly, the separation between the robust embeddings of different types of responses should be maintained. 

Therefore, we define our training loss function as follows:
\begin{align}
        &\mathcal{L}_{\mathrm{ins\_same\_pos}} = z_1*d_1^2 + (1-z_1)*\mathrm{max}(\mathrm{margin}-d_1,0)^2 \nonumber\\ 
        &\mathcal{L}_{\mathrm{ins\_same\_neg}} = z_2*d_2^2 + (1-z_2)*\mathrm{max}(\mathrm{margin}-d_2,0)^2 \nonumber\\ 
        &\mathcal{L}_{\mathrm{out\_robust}} = z_3*d_3^2 + (1-z_3)*\mathrm{max}(\mathrm{margin}-d_3,0)^2 
\end{align}
where $d_i, 1 \leq i \leq 3$ represents the cosine distance and $z_i, 1 \leq i \leq 3$ indicates whether a pair of vectors match; with $z_i=0$ denoting no match and consequently a greater distance between the vectors. Conversely, $z_i=1$ signifies a match leading to a closer distance. To encourage divergence between the robust and non-robust vectors within both positive or negative classes, we set $z_1=z_2=0$. Similarly, for the robust vectors across positive and negative classes to diverge, we assign $z_3=0$.

\begin{align}
    d_1 &= ||h_i^{pr}-h_i^{pn}|| \nonumber\\ 
    d_2 &= ||h_i^{ar}-h_i^{an}|| \nonumber\\ 
    d_3 &= ||h_i^{pr}-h_i^{ar}|| 
\end{align}

In addition, we develop a classification network to classify each factor. The process is defined as follows:
\begin{align}
    h &= \mathrm{concat}(h_c, h_{\mathrm{res}}) \nonumber\\ 
    p_i &= \mathrm{Softmax}(\mathrm{Linear}(h)) \nonumber\\ 
    \mathcal{L}_{cls} &= \sum_{i=0}^2 y_i * p_i
\end{align}
where $h_{res}$ encompasses $\{h_i^{pr}, h_i^{pn}, h_i^{ar}, h_i^{an}\}$, while $y_i=\{0,1,2\}$ is the label.
Specifically, $y_i=1$ corresponds to $h_i^{pr}$, $y_i=0$ relates to $h_i^{ar}$; for all other cases, $y_i=2$.

In summary, the total loss function is
\begin{align}
    \mathcal{L} &= \mathcal{L}_{\mathrm{out}} + \mathcal{L}_{\mathrm{ins\_same\_pos}} + \mathcal{L}_{\mathrm{ins\_same\_neg}} +  \mathcal{L}_{\mathrm{out\_robust}} + \mathcal{L}_{\mathrm{cls}}
\end{align}

\subsection{Evaluation Process}
\subsubsection{SLIDE}
We first follow G-EVAL and prompt LLM to evaluate the given context and response. The prompt could be found in the below:

\textit{
Given a conversation context, which includes 2 speakers[annotated as FS(First Speaker) and SS(Second Speaker)], and a response. \\
Context:\\
{}\\
Response:\\
{}\\
Please finish the following task, do not need to explain it: \\
1. LLM Overall Score: On a continuous scale from 0.0 to 5.0, according to the following criteria to score the context and response.\\
(1)Naturalness: The degree to which a response is naturally written;\\
(2)Coherence: The extent to which the content of the output is well-structured, logical, and meaningful;\\
(3)Engagingness: The degree to which the response is engaging;\\
(4)Groundedness: The extent to which a response is grounded in facts present in the context.\\
}

Thus, we derive a score, denoted as $\mathrm{Score_{LLM}}$. Following this, the SLM model is utilized to encode both the context and the response. In this process, the response encoding is partitioned into two distinct embeddings: robust and non-robust. The formal mathematical representation of this process is as follows:
\begin{align}
    \mathbf{h}_c &= \mathrm{Encoder}(\mathbf{x}_c) \nonumber\\ 
    \mathbf{h}_r &= \mathrm{Encoder}(\mathbf{x}_r) \nonumber\\ 
    \mathbf{h}_{r,\mathrm{robust}}, &\mathbf{h}_{r,\mathrm{non}} = \mathrm{sep}(\mathbf{h}_r)
\end{align}

Then we calculate the cosine distance between context and response, as well as the probability for the given response.
\begin{align}
    d &= \mathrm{cosine\_distance}(\mathbf{h}_c, \mathbf{h}_{r,\mathrm{robust}}) \nonumber\\ 
    s_d &= (d-d_{\mathrm{min}})/(d_{\mathrm{max}}-d_{\mathrm{min}}) \nonumber\\ 
    s_p &= p(y=1) \nonumber\\ 
    &= \mathrm{Softmax}(\mathrm{Linear}(\mathbf{h}_c, \mathbf{h}_{r,\mathrm{robust}})) 
\end{align}
where $d$ is the cosine distance between a context and a response. $d_{min}$ is the smallest distance between context and response for all examples. $d_{max}$ is the biggest distance. The normalized distance is denoted by $s_d$,  while $s_p$ represents the prediction probability of the classifier. 

Empirical results from training process indicate that positive responses are generally closer in embedding space to their respective contexts compared to negative responses, suggesting that $d_{\mathrm{pos}} < d_{\mathrm{neg}}$. Additionally, it has been observed that the value of $s_p$ is typically lower for negative responses than for positive ones. Consequently, it can be inferred that for positive responses, the difference $s_d - s_p$ will be smaller than for negative responses. Building on these observations, we propose a new probabilistic scoring mechanism for our SLM, which is defined as follows:
\begin{align}
\mathrm{Score_{SLM}} = 1 - s_d + s_p
\end{align}
Finally we design an integration strategy to integrate  $\mathrm{Score_{SLM}}$ and $\mathrm{Score_{LLM}}$. Our empirical findings suggest that while the SLM demonstrates greater accuracy in identifying positive responses, the LLMs shows a greater performance on classifying negative ones. 
We define the final score as follows:
\begin{align}
    \mathrm{Score}=\left\{\begin{array}{l}
    \mathrm{Score_{SLM}},\mathrm{Score_{SLM}} \geq 0.5, \\ 
    \mathrm{Score_{LLM}},\mathrm{Score_{LLM}}<0.5,  \\ 
    (\mathrm{Score_{SLM}}+\mathrm{Score_{LLM}})/2,\mathrm{otherwise}
    \end{array}\right.
\end{align}
We set the threshold at 0.5, such that a candidate sample is classified as negative by the SLM if its $\mathrm{Score_{SLM}}$ is below 0.5; otherwise, it is classified as positive.

\subsubsection{DRE}
In contrast to the SLIDE algorithm, DRE relies on a self-learned control loop that allows the framework to find an optimal balance between SLMs and LLMs. As illustrated in Figure \ref{fig:DRE}, the evaluation phase of DRE is primarily divided into two stages: input constrain (interior refinement) and output constrain (exterior refinement).

\begin{figure*}[ht]
\small
\centering 
\includegraphics[scale=0.4]{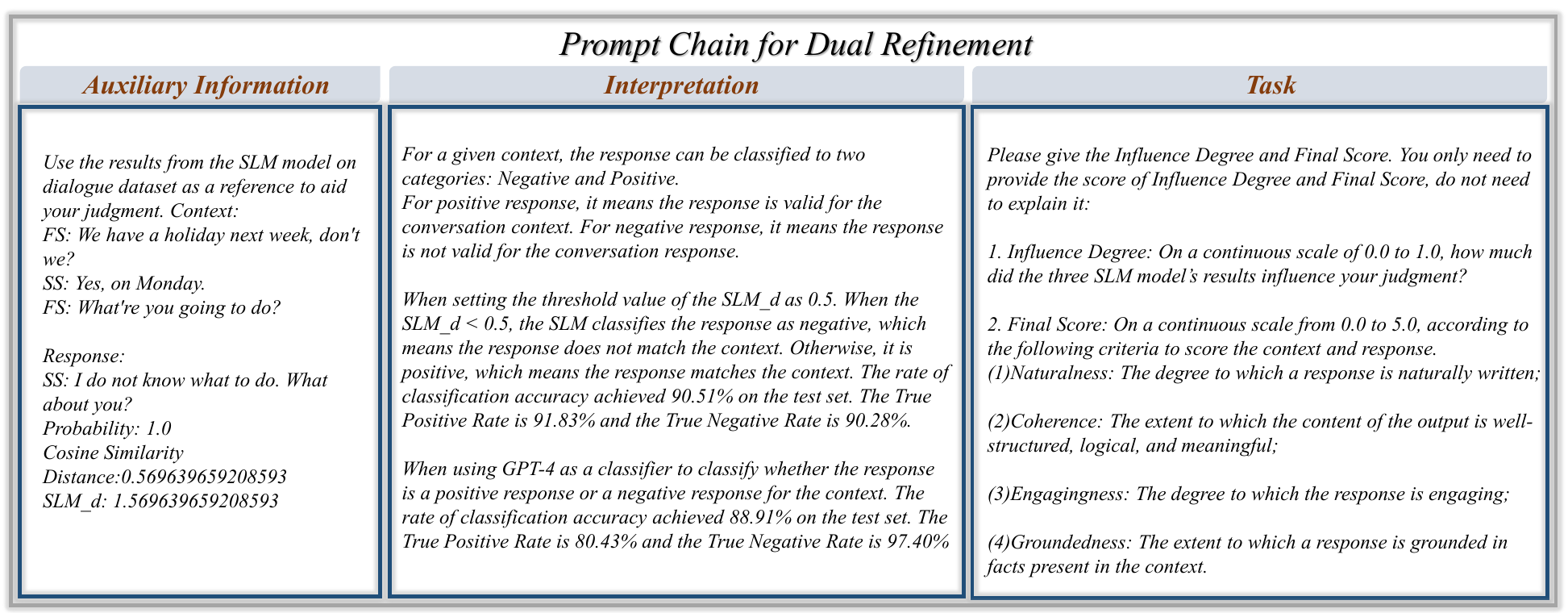}
\caption{The description of prompt. It includes three parts: Auxiliary Information, Interpretation and Task. For the Auxiliary part, it provide the results of the SLM, including probability, cosine similarity and the sum of these two values. For the Interpretation part, we will display the accuracy of classification when using the LLM and the SLM. In the last part, we ask the LLM to conduct two tasks.}
\label{fig:prompt}
\end{figure*}

\paragraph{Interior Refinement} We perform the same initial process as in SLIDE to compute the distance $s_d$ and probability values $s_p$. However, unlike SLIDE, after obtaining these values, we incorporate them into the LLM’s prompts to evaluate the dialogue. The designing of the prompt is shown in the Figure~\ref{fig:prompt}. The Auxiliary Information from the SLM includes the context and corresponding response, along with the previously calculated probability value and distance (as detailed in the preceding section). The Interpretation section defines positive and negative responses and provides the classification accuracy of the SLM and \texttt{GPT-4}. This section aids in interpreting the function of the SLM result, encouraging the LLM to assess it independently rather than directly presenting the classification outcome. This approach enhances the LLM's understanding and utilization of information from the initial prompt. The Task section of the prompt specifies tasks for the LLM, including the Influence degree, which indicates the extent to which the SLM's output influences the evaluation, and the LLM Score. Therefore we obtain two values referred to as $\mathrm{Score_{LLM}}$. 
\begin{align}
    \mathrm{Score}_{\mathrm{LLM}}, s_{Inf} = \mathrm{LLM}(r_i;s_i;t_i)
\end{align} 
We will use these two score from LLM in the next section to calculate the final score.

\paragraph{Exterior Refinement}
In the interior refinement process, we already acquired an initial LLM score $\mathrm{Score_{LLM}}$ and the influence score $\mathrm{Score_{Inf}}$. We assume that this process may also lead to some biases. For example, a positive response can still be given a low score by the LLM. According to this assumption, we could refine the initial LLM score $\mathrm{Score_{LLM}}$ again, which will correct this fault.



We first define $s_c$ as the constraint imposed by the SLM. Then, from the LLM, we derive an influence score $s_{Inf}$, which quantifies the extent to which the SLM influences the LLM. This influence score enables us to constrain the SLM's output in order to evaluate its practical impact on the LLM.  Ultimately, we use the product of $s_c$ and $s_{Inf}$ as our coefficient. Based on these insights, we define a new coeffient for our SLM as follows:

\begin{align}
s_c &= 1 - s_d + s_p \nonumber\\ 
c &= s_c * s_{Inf}
\end{align}

From the the derivation on the previous sections, we understand that the positive response of the $s_c$ is greater than the negative response.  Therefore, positive responses are assigned higher coefficient values, whereas negative responses receive smaller coefficients. Suppose there is a scenario where the LLM assigns a low score to a positive response, while the SLM provides a high coefficient $s_c$, which is greater than 1. To rectify the LLM score, $s_c$ can be utilized to adjust it, resulting in a more accurate and elevated evaluation.

Finally, we use the product of $s_c$ and $s_{Inf}$ to constrain the $Score_{LLM}$. The final score is defined as follows:
\begin{align}
    \mathrm{Score}= s_c * \mathrm{Score_{LLM}}
\end{align}

\section{Experimental Setup}

\subsection{Dataset}
We conduct dialogue evaluation on three open-domain dialogue datasets: DailyDialog++\cite{Sai2020ImprovingDE}, TopicalChat\cite{Gopalakrishnan2019TopicalChatTK}, and PersonaChat~\cite{zhang-etal-2018-personalizing}. The DailyDialog++ dataset contains 9,259 contexts in the training set, 1,028 in the validation set, and 1,142 in the test set. Each context is accompanied by five positive responses, five random negative responses, and five adversarial negative responses. Notably, the DailyDialog++ dataset is the only available dataset that includes multiple human-annotated responses.

In contrast, the PersonaChat and TopicalChat datasets lack multiple positive and adversarial negative responses. To address this limitation, we employ \texttt{GPT-4} \cite{openai2024gpt4technicalreport} to generate both positive and adversarial negative responses and utilize self-refine \cite{madaan2024self} method to modify it. The prompt for generating these responses as follows:

\textit{You are a conversational dialogue generator. \\
Given a conversation context, , which includes 2 speakers[annotated as FS(FirstSpeaker) and SS(SecondSpeaker)], \
and a response.\\
Your task is to generate five diverse positive response and five adversarial negative response respectively. 
\\\\
Positive Response\\
Positive response is valid for the conversation context.\\\\
Adversarial Negative Response\\
Adversarial negative responses have a significant word overlap with the conversation context but are still irrelevant response, which may not have any relation to the context.
You need to choose some words (do not include stopwords such as "I", "you", "are", etc.) from the conversation context and use them directly or indirectly while writing the adversarial negative responses. 
Indirect usage here refers to using words closely related to the context words. \\
For example,using synonyms,antonyms, homonyms, subwords, or other words that are known to frequently co-occur with the words in the context (e.g., the words  "flexibility" and "injuries" co-occur with "acrobatics").
\\\\
The following are five examples of a conversation context and response, and the corresponding prediction. \\\\
Example\\
Context: \\
FS:Is there something wrong?\\
SSI enjoy having your daughter in my class.\\
FS:I'm glad to hear it.\\
Positive response: 
She is so brilliant.\\
Her behavior is good in the class.\\
I would love to hear that she knows every rules and regulation.\\
I was shocked to know that she is your daughter.\\
She answers all my questions.\\\\
Adversarial Negative Responses:\\
I enjoy listening jazz music in my free time.\\
I need pin drop silence in the class. \\
If I hear someone talking they will be sent out of the class.\\
I am glad you enjoyed the magic show organised by our team. \\
I think there was something wrong with the CCTV camera installed in the class.\\
This is the wrong method to solve the problem. Please be attentive in the class.\\}

Consequently, the generated training and test sets for both the PersonaChat and TopicalChat datasets comprise $2,000$ contexts each. Each context within these datasets is further enriched with a set of responses, consisting of five positive responses and five adversarial negative responses. We randomly selected $1,200$ generated responses from the test set for evaluation by human annotators. A significant majority, over $98\%$, of these responses were deemed appropriate for their contexts. The overall generation algorithm is shown as shown in the Algorithm~\ref{alg:dataset}. The statistics of the final datasets are shown in the Table~\ref{table:dd++}. Table~\ref{tab:examples} has shown some examples of the datasets.

The training sets of the three datasets are used to train the SLM models. For evaluation, we randomly sample a subset of examples from the test sets. Native speakers are then asked to revise each sample to improve fluency and enhance human-likeness, thereby mitigating potential self-bias. Finally, human annotators are invited to score the revised samples.

\begin{table}[h]
\centering
\begin{tabular}{llll}
\toprule
\multicolumn{1}{l}{\textbf{Context:}}& \multicolumn{3}{l}{\begin{tabular}{p{4.5cm}}
Hi how are you doing? pretty good . Just finished up an art project i've been working on. That is nice are you an artist \end{tabular}}\\ \hline 
\multicolumn{1}{l}{\textbf{Positive Response:}} & \multicolumn{3}{l}{\begin{tabular}{p{4.55cm}}
I wouldn't call myself an artist, but I enjoy creating art.\end{tabular}}\\  \hline 
\multicolumn{1}{l}{\textbf{Adversarial Response:}} & \multicolumn{3}{l}{\begin{tabular}{p{4.5cm}}
Finished up with the new book by my favorite author.\end{tabular}}\\  \hline 

\multicolumn{1}{l}{\textbf{Context:}} & \multicolumn{3}{l}{\begin{tabular}{p{4.5cm}}Did you see the movie Black Panther? I did see it. Did you see it? Yeah! One of the best movies in 2018 I think.  What did you think?
\end{tabular}} \\ \hline
\multicolumn{1}{l}{\textbf{Positive Response:}} & \multicolumn{3}{l}{\begin{tabular}{p{4.5cm}}It really was one of the standout films of that year.\end{tabular}} \\ \hline
\multicolumn{1}{l}{\textbf{Adversarial Response:}} & \multicolumn{3}{l}{\begin{tabular}{p{4.5cm}}
One of the best pizzas I had was in 2018 at this small Italian place downtown.\end{tabular}}\\  \hline 

\multicolumn{1}{l}{\textbf{Context:}}& \multicolumn{3}{l}{\begin{tabular}{p{4.5cm}}Hi, how are you? I am well, thank you. Did you know that art, music and poetry were once olympic sports?? I did not. I did know there is a 224 word palindrome. \end{tabular}}\\\hline 
\multicolumn{1}{l}{\textbf{Positive Response:}} & \multicolumn{3}{l}{\begin{tabular}{p{4.5cm}}That's fascinating about the palindrome! I love learning new things.\end{tabular}}             \\ \hline 
\multicolumn{1}{l}{\textbf{Adversarial Response:}} & \multicolumn{3}{l}{\begin{tabular}{p{4.5cm}}
Did you know that many famous musicians started their careers in gospel choirs?\end{tabular}}\\  
\bottomrule
\end{tabular}

\caption{Samples of context–response pairs from three datasets. Each example includes a context, a positive response, and an adversarial response. The positive response is semantically aligned with the context, while the adversarial response is not, although it may share some similar words with the context.}
\label{tab:examples}
\end{table}

\begin{algorithm}[t]

\caption{Dataset Generation and Refinement}
\label{alg:dataset}
\begin{algorithmic}[1]
\REQUIRE A set of $n$ data items $D = \{d_1, d_2, \ldots, d_n\}$
\ENSURE Generate five positive responses and five adversarial negative responses $P = \{p_1, t_2, \ldots, p_5\}$ and $A = \{a_1, a_2, \ldots, a_5\}$ where each $p_i$ is a valid response  of $d_i$ and $a_i$ is a negative response  of $d_i$
\FOR{$i = 1$ to $n$}
    \REPEAT
        \STATE $<p_i, a_i> \leftarrow \text{LLM-Generate}(d_i)$
        \STATE $correct \leftarrow \text{LLM-Check}(d_i, p_i) and (d_i, a_i)$
    \UNTIL{$correct$}
\ENDFOR
\STATE \textbf{Sample examples and human check}
\STATE \textbf{SLM CLASSIFICATION}
\STATE \textbf{return} $T$
\end{algorithmic}
\end{algorithm}

\begin{table}[ht]
\scriptsize
\centering
\resizebox{1.0\textwidth}{!}{
\begin{tabular}{l|c|c|c}
\toprule
\textbf{Datasets} & \textbf{Train} & \textbf{Val} & \textbf{Test} \\ \midrule
\multicolumn{4}{l}{\textbf{DailyDialog++}} \\
\midrule
Contexts                      & 9,259           &1,028 & 1,142          \\ 
Positive responses           & 46,295          &5,140 & 5,710          \\ 
Adversarial negative responses & 46,295         &5,140 & 5,710  \\ 
\midrule
\multicolumn{4}{l}{\textbf{TopicalChat}} \\
\midrule
Contexts & 8,628	& 539	& 539 \\
Positive responses & 16,678 & 2,695 & 2,695 \\
Adversarial negative responses & 8,050 & 2,695 & 2,695 \\
\midrule
\multicolumn{4}{l}{\textbf{PersonaChat}} \\
\midrule
Contexts & 10,907	& 1,000	& 968 \\
Positive responses & 18,957 & 5,000 & 4,840 \\
Adversarial negative responses & 8,050 & 5,000 & 4,840 \\
\bottomrule
\end{tabular}}
\caption{Statistics of the datasets. Each context has 5 positive responses and 5 adversarial negative responses in each dataset.}
\label{table:dd++}
\end{table}

\subsection{Baselines}
We adopted a diverse array of baseline metrics to comprehensively evaluate dialogue quality in our study. Among the word-overlap and embedding-based metrics, we included well-established benchmarks such as BLEU \cite{papineni-etal-2002-bleu}, ROUGE \cite{lin-2004-rouge}, METEOR \cite{banerjee-lavie-2005-meteor}, Embedding-Average \cite{Wieting2016TowardsUP}, Vector-Extrema \cite{forgues2014bootstrapping}, BERTScore \cite{Zhang2019BERTScoreET}, Unieval \cite{zhong-etal-2022-towards}, BARTScore \cite{yuan2021bartscore} and CMN \cite{zhao2023evaluating}. These metrics are widely recognized for their ability to assess the linguistic quality and semantic similarity of generated text. Additionally, for evaluating LLM-based metrics, we incorporated recent advancements in the field, specifically G-EVAL \cite{liu-etal-2023-g} and an LLM-based metric proposed by \cite{Chiang2023CanLL}, which we refer to as LLM-Chiang. These LLM-based metrics provide insights into the performance of large-scale language models specifically tailored for dialogue evaluation tasks, complementing the traditional word-overlap and embedding-based metrics with their unique capabilities.

\subsection{Evaluation Set}
For the evaluation set, we selected 600 context-response pairs from each of three open-domain dialogue datasets—DailyDialog++, PersonaChat, and TopicalChat—a total of $1,800$ samples. The evaluation was conducted by three human annotators, all proficient in English, to ensure accurate assessments. The annotators were instructed to thoroughly review each context-response pair and rate them according to four distinct criteria: naturalness, coherence, engagingness, and groundedness, utilizing a 1-5 Likert scale for their evaluations.

The four criteria used in our evaluation framework are aligned with those established by ~\cite{zhong-etal-2022-towards} and are crucial for assessing the quality of dialogue systems. First, Naturalness refers to how closely a generated response resembles natural human language, emphasizing fluency and grammatical correctness. Second, Coherence measures the logical flow and structure of the response, ensuring that it is cohesive and maintains a clear narrative or argument. Third, Engagingness evaluates the extent to which the response captivates and maintains the interest of the interlocutor, considering factors like creativity, empathy, and the ability to sustain conversation. Lastly, Groundedness assesses how well the response is supported by factual information present in the dialogue context or external knowledge sources, ensuring accuracy and relevance to the conversation. These criteria collectively provide a comprehensive framework for evaluating the effectiveness and quality of dialogue generation models across various contexts and datasets.

Upon obtaining scores for each criterion, an aggregate score was calculated by averaging across all criteria to provide a comprehensive measure of each response's quality. Additionally, the Inner-Annotator Agreement (IAA) among the three evaluators was assessed using Cohen's Kappa~\cite{doi:10.1177/001316446002000104}, which indicated a substantial agreement level (0.53-0.80) among annotators, with an average value of 0.53, thus validating the reliability of the human evaluation.

For the evaluation task in our study, fairness is a crucial consideration. As we generated responses using \texttt{GPT-4-1106} \cite{achiam2023gpt} on the PersonaChat and TopicalChat datasets, so we avoid to evaluate these responses using \texttt{GPT-4} again as it could introduce self-bias and lead to an unfair comparison due to the model's familiarity with its own generated responses. To mitigate this potential issue and ensure a more objective evaluation, we opted to use alternative models for evaluation purposes. Specifically, we employed \texttt{GPT-3.5-turbo-1106}, \texttt{Claude-3-haiku-20240307}, and \texttt{Gemini-1.5-flash} \cite{team2023gemini} for evaluating the generated responses, as well as some open-sourced LLMs (\texttt{Llama3.3-70B} \cite{touvron2023llama}, \texttt{Qwen2.5-70B} \cite{bai2023qwen}, and \texttt{DeepSeek-R1-70B} \cite{guo2025deepseek}). By using different models for evaluation that were not involved in the generation process, we aimed to provide a more unbiased assessment of dialogue quality across different datasets and conditions. Additionally, we have recruited native speakers to revise each sample in the evaluation set to make it more human-like. These two approaches help to ensure that our evaluation results are more robust and reflective of the models' generalization capabilities beyond the training and generation phase. 

\subsection{Computational Cost}

Regarding computational costs, the bulk of our expenses—approximately $1,000$ dollars—was incurred during response generation. To facilitate reproducibility, all datasets are publicly available on \texttt{HuggingFace}. 

Evaluation expenses were kept to around $100$ dollars for \texttt{GPT-3.5} and \texttt{Claude-3} via their APIs, while \texttt{Gemini’s} free API required only time. For \texttt{Llama3.3-70B}, \texttt{Qwen2.5-70B}, and \texttt{DeepSeek-R1-70B}, we performed evaluations locally using Ollama on an RTX A6000 GPU, each model occupying roughly 40 GB of GPU memory in inference mode. In terms of speed, \texttt{Llama3.3-70B} and \texttt{Qwen2.5-70B} generated both the influence and LLM scores in about one minute per example, whereas DeepSeek took approximately three minutes per sample.

For training our secondary language model, we used the Sentence-Transformer framework with the \texttt{DistilBERT-base-uncased} backbone (66 M parameters). We fine-tuned it on an A6000 GPU with a batch size of 32, initializing from the pretrained weights available on \texttt{HuggingFace}. The training process took almost 3 hours.

\section{Experimental Results}

\subsection{Analysis of Training Process}

\begin{figure*}[htb]
\small
\centering 
\includegraphics[scale=0.4]{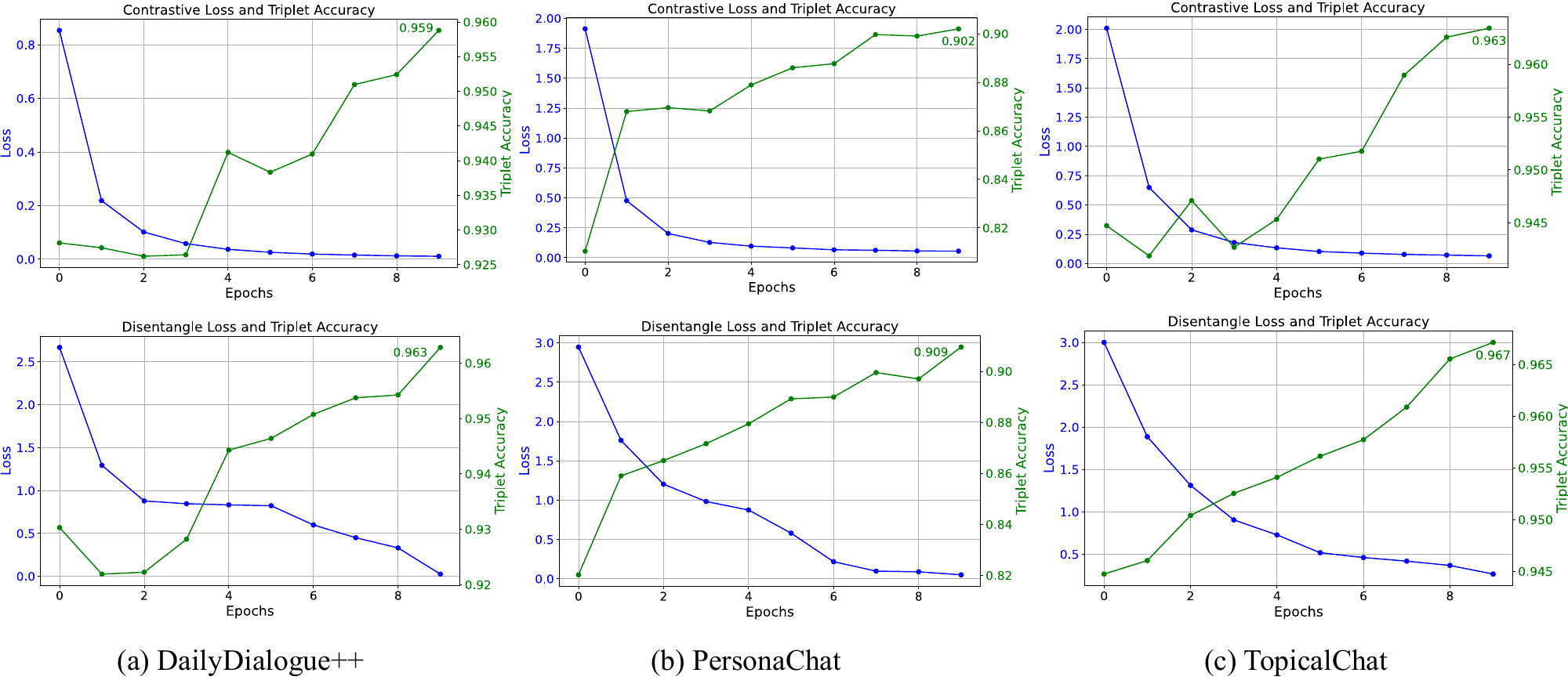}
\caption{Contrastive Loss and Triplet Accuracy of SLM. The blue line shows the contrastive loss, which decreases as the number of epochs increases, while the green line represents the triplet accuracy, which improves with each additional epoch.}
\label{fig:contrastive}
\end{figure*}

\begin{figure*}[ht]
\centering 
\includegraphics[scale=0.4]{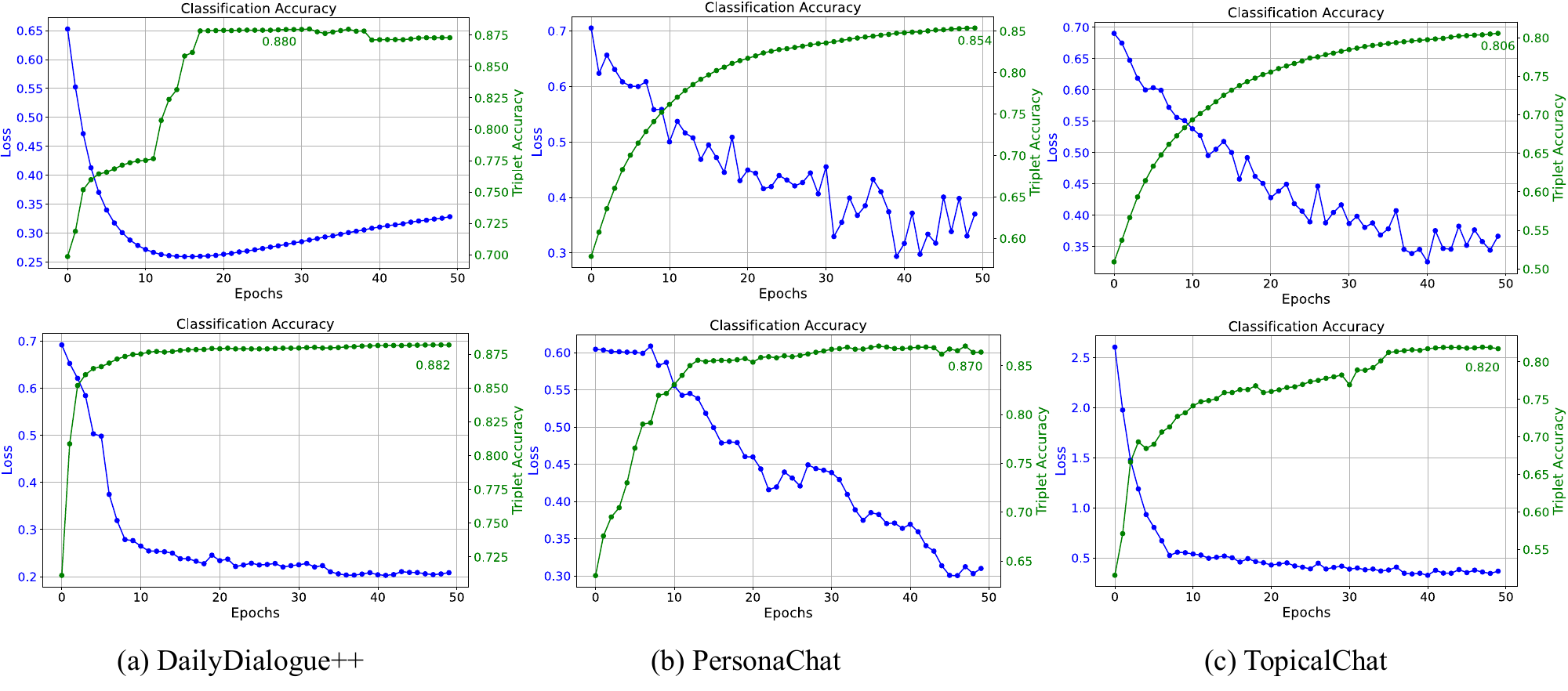}
\caption{Classification Loss and  Accuracy SLM. The blue line shows the classification loss, which decreases as the number of epochs increases, while the green line represents the classification accuracy, which improves with each additional epoch.}
\label{fig:class}
\end{figure*}
To demonstrate the effectiveness of the disentanglement process, we analyze the changes in loss and accuracy values across three datasets.  This includes an analysis of contrastive loss and triplet accuracy, which measures the proportion of instances where the distance between positive responses and their corresponding context is smaller than that of negative responses, as well as the classification rate and its associated accuracy.  Figure ~\ref{fig:contrastive}  illustrates that models incorporating the disentanglement process achieve higher triplet accuracy, indicating that they are more effective at positioning positive responses closer to their context compared to models without disentanglement. Additionally, we employed these models to encode both the context and responses, followed by training a classification model, where the variation in loss and accuracy across epochs was visualized, as shown in Figure ~\ref{fig:class}. In this figure, the top panels depict the classification process for models without disentanglement, while the bottom panels represent the same process for models with disentanglement across the three datasets. The results clearly indicate that the accuracy of the disentangled models surpasses that of the non-disentangled models. These findings demonstrate that the disentanglement process offers superior performance compared to the contrastive-only approach.

\begin{figure*}[htb]
\small
\centering 
\includegraphics[scale=0.4]{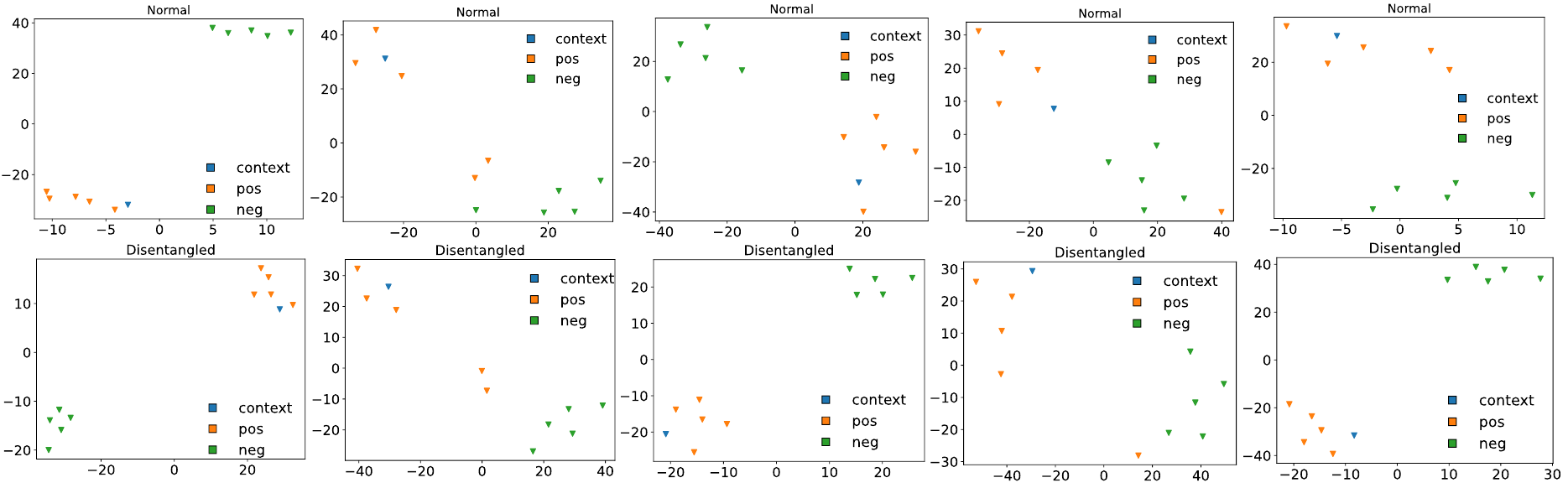}
\caption{T-SNE visualization of sentence representations for context and responses across several examples. Each example includes one context, five positive responses, and five adversarial negative responses. The top row represents the vector space before disentanglement, labeled as "Normal," while the bottom row shows the vector space after disentanglement, labeled as "Disentangled." The visualizations illustrate that, following the disentanglement process, positive responses (in orange) are positioned closer to their corresponding contexts (in blue) compared to the negative responses (in green), demonstrating the improved clustering effect of the disentanglement process.}
\label{fig:label1}
\end{figure*}

To further demonstrate the effectiveness of the disentanglement process, we selected a subset of examples from the DailyDialog++ dataset and used T-SNE for visualization, as shown in Figure~\ref{fig:label1}. The vector representations prior to disentanglement are shown in the top row, while the post-disentanglement representations are presented in the bottom row. The visualizations after disentanglement demonstrate a more distinct clustering of positive responses in proximity to their corresponding contexts. These results provide additional confirmation of the disentanglement process's effectiveness.

\subsection{Analysis for Different Thresholds of SLM}

\begin{figure}[ht]
\centering 
\includegraphics[scale=0.4]{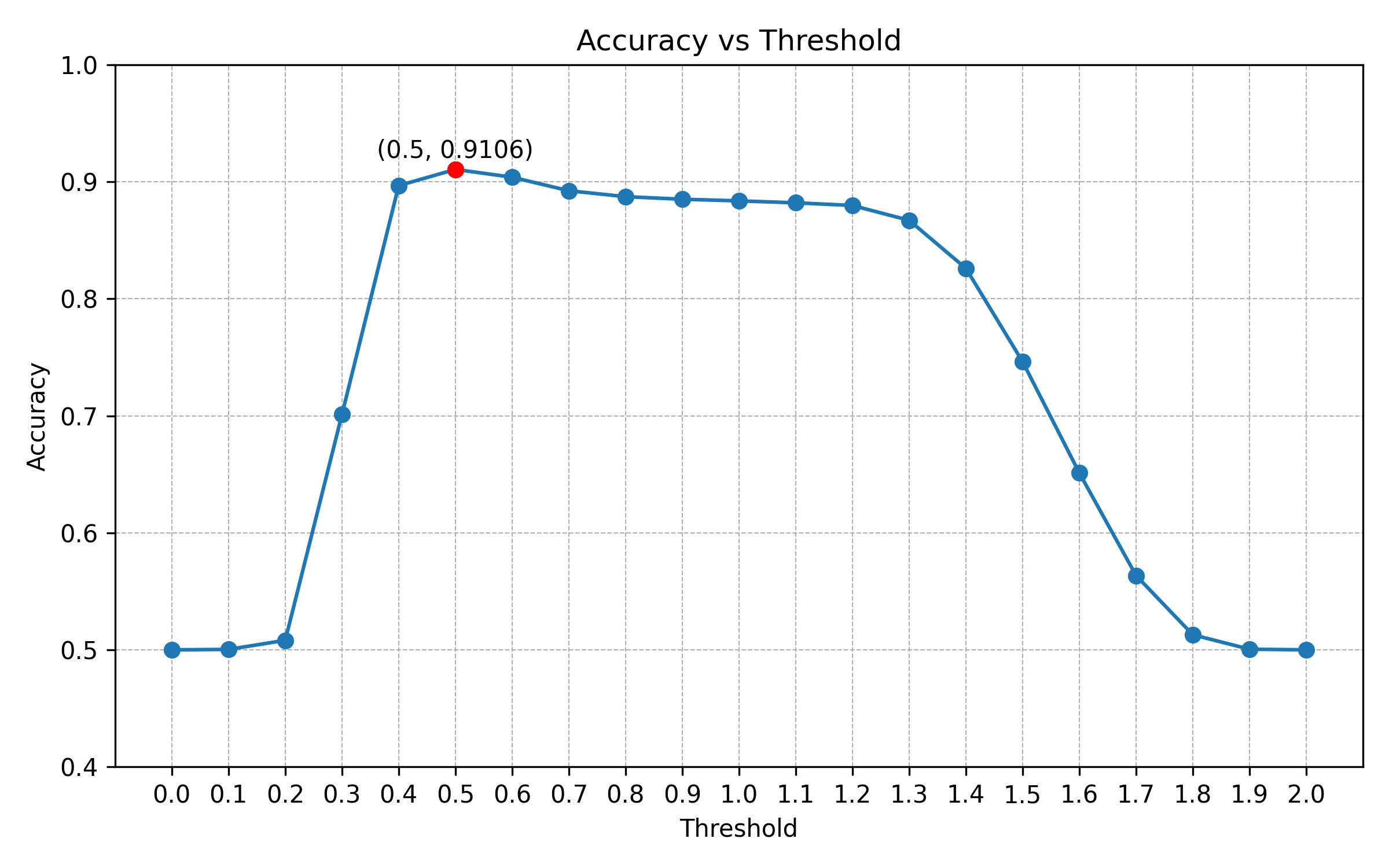}
\caption{Classification Accuracy with Different Thresholds in the DailyDialogue++ Dataset.}
\label{fig:acc with diff threshold}
\end{figure}

For the SLIDE Algorithm, we have set a threshold 0.5 for the SLM score, which means that when the
$Score_{SLM}$ < 0.5, the SLM classifies the response as negative, which
means the response does not match the context. Otherwise, it is
positive, which means the response matches the context. To analyze how different thresholds affect accuracy, we conducted experiments using multiple threshold values and found that 0.5 yielded the most consistent and optimal performance. As shown in Figure \ref{fig:acc with diff threshold}, the threshold of 0.5 achieved the highest accuracy.

\subsection{Dialogue Classification Task}
We conduct dialogue classification on the on DailyDialog++ dataset as it contains the human-annotated labels for each context-response pair.
We calculate the classification accuracy using SLM and LLMs (i.e., \texttt{GPT-3.5} and \texttt{GPT-4}). 
We also use SLM that trained on the identical dataset for the ablation study. 

\begin{table}[ht]
\small
\centering
\resizebox{0.9\linewidth}{!}{
\begin{tabular}{l|c|c|c}
\toprule
\textbf{Model} & \multicolumn{1}{l|}{\textbf{Positive}} & \multicolumn{1}{l|}{\textbf{Negative}} & \multicolumn{1}{l}{\textbf{Overall}} \\
\midrule
\textbf{\texttt{GPT-3.5}}& 59.36 & 91.01 & 75.18\\
\textbf{\texttt{GPT-4}} & 80.43 & \textbf{97.40} & 88.91 \\
\textbf{\texttt{Gemini}} & 80.62 & 95.13 & 87.86 \\
\textbf{\texttt{Claude3}} & 80.77 & 96.05 & 88.41 \\
\textbf{SLM (Dis)} & 79.35  & 90.03 & 84.00  \\ 
\textbf{SLM (Prob)} & 83.25 & 93.11& 88.19 \\ 
\textbf{SLM (Prob\& Dis)} & \textbf{91.83}& 90.28 & \textbf{91.05} \\
\bottomrule
\end{tabular}
}
\caption{Classification accuracy for only the Positive and adversarial negative responses, and the overall accuracy. Dis and Prob refers to the distance measures and classification probability, respectively.}
\label{tab:classification}
\end{table}

As Table \ref{tab:classification} shows, SLM employs both probability and distance measures, achieves superior performance with an accuracy of 91.05\%, which exceeds the 88.91\% accuracy achieved by \texttt{GPT-4}. 
Experimental results show that the SLM  particularly excels in the classification of positive responses, achieving the highest accuracy of 91.83\%. 
Conversely, \texttt{GPT-4} exhibits the strongest performance in classifying adversarial negative responses,  achieving 97.40\% accuracy. The performance of \texttt{Gemini} surpasses that of \texttt{GPT-3.5}, yet remains inferior to that of \texttt{GPT-4}. These results indicate that while LLMs, such as \texttt{GPT-4}, possess robust capabilities in recognising adversarial negative responses, they still struggle with the one-to-many problem, especially when dealing with semantically diverse positive responses. 
On the other hand, SLM with the integration of distance and probability metrics shows exceptional performance in recognising open-domain  positive responses. Based on these insights, we propose a hybrid approach that combines the SLM, a task-specific model, with LLMs for comprehensive dialogue evaluation. 
This approach is designed to leverage the distinct strengths of both types of models to enhance the overall performance in open-domain dialogue evaluations.

\subsection{Dialogue Evaluation}
\begin{table*}[htb]
\centering

\resizebox{0.85\linewidth}{!}{
\scriptsize
\begin{tabular}{lcc|cc|cc}
             \toprule
& \multicolumn{2}{c|}{DailyDialog++} & \multicolumn{2}{c|}{PersonaChat} & \multicolumn{2}{c}{TopicalChat}\\ \midrule
Metrics & Pearson's $\rho$ & Spearman's $\tau$ & Pearson's $\rho$ & Spearman's $\tau$ & Pearson's $\rho$ & Spearman's $\tau$ \\ \midrule

BLEU-1 & 0.303  & 0.242  & 0.331  & 0.276  & 0.352  & 0.330 
        \\
BLEU-2 & 0.317  & 0.270  & 0.311  & 0.261  & 0.279  & 0.252 
        \\
BLEU-3 & 0.256  & 0.251  & 0.251  & 0.239  & 0.179  & 0.189 
        \\
BLEU-4 & 0.234  & 0.247  & 0.214  & 0.217  & 0.247  & 0.154 
        \\ \midrule

ROUGE-1 & 0.303  & 0.270  & 0.304  & 0.250  & 0.374  & 0.356 
\\
ROUGE-2 & 0.178  & 0.133  & 0.142  & 0.126  & 0.173  & 0.149 
        \\
ROUGE-L & 0.305  & 0.271  & 0.294  & 0.244  & 0.279  & 0.259 
        \\ \midrule

METEOR & 0.196  & 0.142  & 0.250  & 0.200  & 0.293  & 0.278 
        \\ \midrule[1pt]
Embedding 
        \\
Extrema & 0.382  & 0.367  & 0.330  & 0.299  & 0.389  & 0.244 
    \\
Greedy & 0.322  & 0.272  & 0.328  & 0.290  & 0.272  & 0.248 
    \\
Average & 0.186  & 0.208  & 0.239  & 0.249  & 0.254  & 0.261 
    \\ \midrule
BERTScore & 0.419  & 0.381  & 0.465  & 0.412  & 0.525  & 0.485  \\

BARTSCORE & 0.305  & 0.264  & 0.466  & 0.442  & 0.470  & 0.427 
\\
Unieval         & 0.103  & 0.069  & 0.192  & 0.131  & 0.357  & 0.290              \\

CMN             & 0.448  & 0.397  & 0.502  & 0.482  & 0.546  & 0.490              \\

\midrule[1pt]

G-EVAL (\texttt{GPT-3.5}) & 0.447  & 0.414  & 0.516  & 0.562  & 0.642  & 0.668  \\

G-EVAL (\texttt{Gemini}) & 0.511  & 0.504  & 0.552  & 0.544  & 0.611  & 0.605  \\

G-EVAL (\texttt{Claude3}) & 0.527  & 0.544  & 0.569  & 0.559  & 0.621  & 0.630  \\ \midrule

LLM-Chiang (\texttt{GPT-3.5}) & 0.632  & 0.583  & 0.599  & 0.670  & 0.667  & 0.661  \\

LLM-Chiang (\texttt{Gemini}) & 0.622  & 0.601  & 0.597  & 0.669  & 0.677  & 0.652  \\

LLM-Chiang (\texttt{Claude3}) & 0.644  & 0.611  & 0.620  & 0.678  & 0.670  & 0.669  \\

\midrule[1pt]






SLIDE & & & & \\

Ours (SLM-Prob-only) & 0.677 & 0.632 & 0.638  & 0.692  & 0.691  & 0.689  \\

Ours (SLM-Dis-only) & 0.652  & 0.600  & 0.668  & 0.692  & 0.642  & 0.673  \\

Ours (SLM-Dis and Prob) & 0.728  & 0.666  & 0.711  & 0.707  & 0.741  & 0.727  \\ 

Ours (LLMs-only (\texttt{GPT-3.5})) & 0.518  & 0.472  & 0.506  & 0.617  & 0.590  & 0.669  \\

SLIDE (\texttt{GPT-3.5})) & 0.727  & 0.674  & 0.692  & 0.695  & 0.696  & 0.718  \\ 

Ours (LLMs-only (\texttt{Gemini})) & 0.660  & 0.614  & 0.719  & 0.688  & 0.701  & 0.699 \\ 

SLIDE (\texttt{Gemini}) & \textbf{0.760 } & 0.689  & 0.701  & 0.705  & 0.713  & 0.712  \\

Ours (LLMs-only (\texttt{Claude3})) & 0.674  & 0.622 & 0.721  & 0.693  & 0.704  & 0.702  \\

SLIDE (\texttt{Claude3}) & 0.688  & 0.699  & 0.712  & 0.719  & 0.721  & 0.725  \\
\midrule[1pt]
DRE & & & & \\

Non-DRE(\texttt{GPT-3.5}) & 0.448  & 0.415  & 0.606  & 0.550  & 0.643  & 0.668  \\

In-DRE(\texttt{GPT-3.5}) & 0.459  & 0.537  & 0.472  & 0.562  & 0.624  & 0.688  \\

Ex-DRE(\texttt{GPT-3.5}) & 0.723  & 0.671  & 0.706  & 0.731  & 0.725  & 0.740  \\

DRE(\texttt{GPT-3.5}) & \textbf{0.744 } & \textbf{0.688 } & \textbf{0.729 } & \textbf{0.748 } & \textbf{0.742 } & \textbf{0.755 } \\ \midrule

Non-DRE(\texttt{Claude3}) & 0.670  & 0.620  & 0.629  & 0.617  & 0.638  & 0.642  \\

In-DRE(\texttt{Claude3}) & 0.617  & 0.582  & 0.585  & 0.656  & 0.614  & 0.626  \\

Ex-DRE(\texttt{Claude3}) & 0.742  & 0.698  & 0.724  & 0.728 & 0.732  & 0.717  \\

DRE(\texttt{Claude3}) & \textbf{0.753 } & \textbf{0.703 } & \textbf{0.730 } & \textbf{0.737} & \textbf{0.747 } & \textbf{0.739 } \\ \midrule

Non-DRE(\texttt{Gemini}) & 0.615  & 0.568  & 0.703  & 0.652  & 0.611  & 0.671  \\

In-DRE(\texttt{Gemini}) & 0.574  & 0.541  & 0.671  & 0.631  & 0.606  & 0.667  \\

Ex-DRE(\texttt{Gemini}) & 0.717  & 0.677  & 0.732  & 0.743  & 0.660  & 0.684  \\

DRE(\texttt{Gemini}) & \textbf{0.748 } & \textbf{0.695 } &\textbf{ 0.752 } & \textbf{0.748 } & \textbf{0.726 } & \textbf{0.737 } \\ \midrule

Non-DRE(\texttt{llama3.3-70B}) &0.635  &0.648   &0.609  &0.617  &0.642  &0.631  \\

In-DRE(\texttt{llama3.3-70B}) &0.685   &0.667  &0.613  &0.657   &0.664   &0.638   \\

Ex-DRE(\texttt{llama3.3-70B}) &0.677   &0.650   &0.607   &0.637   &0.652   &0.634   \\

DRE(\texttt{llama3.3-70B}) &\textbf{0.699}   &\textbf{0.687}   &\textbf{0.616}  &\textbf{0.687}   &\textbf{0.674}   &\textbf{0.643}   \\ \midrule

Non-DRE(\texttt{Qwen2.5-72B}) &0.648   &0.631   &0.559   &0.624   &0.551   &0.517   \\

In-DRE(\texttt{Qwen2.5-72B}) &0.668   &0.663   &0.572   &0.659   &0.599   &0.579   \\

Ex-DRE(\texttt{Qwen2.5-72B}) &0.657   &0.641   &0.666  &0.610  &0.621  &0.554  \\

DRE(\texttt{Qwen2.5-72B}) &\textbf{0.694}  &\textbf{0.683}   &\textbf{0.686}  &\textbf{0.673}  &\textbf{0.648}   &\textbf{0.632}  \\ \midrule

Non-DRE(\texttt{DeepSeek-R1-70B}) &0.670  &0.584  &0.662  &0.637  &0.636  &0.666  \\

In-DRE(\texttt{DeepSeek-R1-70B}) &0.673  &0.579  &0.667  &0.672  &0.626  &0.711   \\

Ex-DRE(\texttt{DeepSeek-R1-70B}) &0.643   &0.560  &0.646  &0.658  &0.657   &0.677  \\

DRE(\texttt{DeepSeek-R1-70B}) &\textbf{0.703}  &\textbf{0.647}   &\textbf{0.705}  &\textbf{0.701}  &\textbf{0.671}  &\textbf{0.718}  \\ 

\bottomrule

\end{tabular}
}
\caption{Pearson and Spearman correlations with human judgements. Dis and Prob refers to the distance measures and classification probability, respectively. For DRE, we did some ablation study. Non-DRE involved the LLM operating without the assistance of the SLM. In-DRE involved the LLM using the results of the SLM but without refinement by the coefficient $c$. Ex-DRE means the LLM did not utilize the SLM results but underwent refinement by $c$. 
}
\label{tab:similar_correlation}
\end{table*}
We firstly compared the SLIDE with baselines. As shown in Table \ref{tab:similar_correlation}, we investigate the correlation between automatic evaluation metrics and human judgments across three datasets. The word-overlap metrics, including BLEU, ROUGE, and METEOR, exhibit a weak positive correlation with human scores ($<$0.4), among which ROUGE-2 showing the least correlation ($<$0.2). Similarly, embedding-based metrics, such as Embedding and BARTScore,  demonstrate a weak correlation with human evaluations, ranging from 0.1 to 0.4. Notably, CMN shows the strongest correlation to human judgments among both word-overlap and embedding based traditional metrics. On the other hand, Unieval achieves the suboptimal results ($<$0.2) on the DailyDialog++ and PersonaChat datasets .

As for LLM-Based metrics, there is a notable improvement in correlation with human scores compared with traditional metrics, most show correlations greater than 0.6. Generally, LLM-Chiang exhibits superior performance compared with G-EVAL. Specifically, LLM-Chiang (\texttt{Claude3}) achieves the highest correlation in all three datasets, except for LLM-Chiang (\texttt{Gemini}) has achieved the highest Pearson correlation (0.677) in TopicalChat dataset.
Moreover, 
the variance in performance between G-EVAL and LLM-Chiang indicates the significant impact of model selection (i.e., \texttt{Gemini} or \texttt{GPT-3.5}) and prompt design on LLM-based metric effectiveness.

As for the SLM, Our SLM-Dis and Prob approach that incorporate both distance and probability measures, revealing a strong correlation with human evaluations ($>$0.6). 
As for the SLIDE method, it combines an SLM with LLMs. SLIDE (\texttt{Gemini}) achieves the strongest Pearson correlation with human ratings on the DailyDialog++ dataset (0.760) and SLIDE (\texttt{Claude3}) achieve the best Spearman correlation on it (0.699). For other two datasets, the SLIDE (Claude3) achieves the best results (0.712 (Pearson) and 0.719 (Spearman) in the PersonaChat, 0.721 (Pearson) and 0.725 (Spearman) in the TopicalChat). 
Furthermore, the amalgamation of LLM and SLM techniques results in a significant enhancement in correlation strength, surpassing both G-EVAL and LLM-Chiang, as well as Ours (LLM-only). The same trend is also observed when utilizing \texttt{Gemini} and \texttt{Claude3}. These three distinct models could demonstrate the effectiveness of integrating LLM and SLM techniques.


To validate the effectiveness of distance metric used in SLM, we conduct several ablation studies. 
It can be observed that Ours (SLM-Dis-Prob) gives better performance than
the model variant with the Dis (SLM-Dis-only) or Prob (SLM-Prob-only) component only.
This suggests that SLM with distance metric is more effective in
evaluating open-domain dialogues.

However, the SLIDE is only a simple combination between SLM and LLM. We continue to see the performance of DRE. To demonstrate the effectiveness of the dual-refined method, we designed a series of ablation experiments. The first experiment involved the LLM operating without the assistance of the SLM, namely Non-DRE. The second experiment involved the LLM using the results of the SLM but without refinement by the coefficient $c$
 , referred to as In-DRE. In the third ablation experiment, the LLM did not utilize the SLM results but underwent refinement by $c$, which we denote as Ex-DRE.


Specifically, in the DailyDialog++ dataset, \texttt{Claude3} performed the best, achieving 0.753 in Pearson correlation and 0.703 in Spearman correlation. \texttt{Gemini} achieved the best results in the PersonaChat dataset (0.752 in Pearson and 0.748 in Spearman). For the TopicalChat dataset, \texttt{Claude3} performed best in Pearson correlation (0.747), while \texttt{GPT-3.5} achieved the highest Spearman correlation (0.755).

In the ablation studies, we observed that the interior refinement method (In-DRE) generally exhibited inferior performance compared to the baseline method (Non-DRE) across three different LLMs. However, when only exterior refinement (Ex-DRE) was used, the correlation with human scores increased significantly compared to both Non-DRE and In-DRE, with an increase of approximately 0.1-0.3. This phenomenon indicates that exterior refinement is more effective than the interior refinement and leads to better evaluation scores. Additionally, although only applying interior refinement  may have a worse impact, the model achieved the best performance across all three LLMs when combining both the interior and exterior refinement approaches. This result suggests that the interior refinement plays an auxiliary role in assisting exterior refinement to finely adjust the model into the more bias-free judgement. Solely using interior refinement is too minor to affect LLMs leveraging the constraints represented by SLMs, leading to an converse impact. These findings also indicate that our dual-refinement method is suitable for different LLMs and can significantly enhance their evaluation performance in open-domain dialogue tasks. The superior performance achieved by applying our dual-refinement to any of the selected LLMs also further demonstrates that task-specific knowledge could alleviate the hallucination issues thus leading to a better correlation with human scores.

We further deploy three middle, open-source LLMs (\texttt{Llama3.3-70B}, \texttt{Qwen2.5-70B}, and \texttt{DeepSeek-R1-70B}) as evaluators, our DRE method achieved superior performance across in these models. This confirms the method’s generalizability across diverse LLM architectures. Notably, exterior refinement did not universally dominate performance gains; in some cases, interior refinement exerted a greater impact, underscoring the necessity of integrating both stages for optimal results.

Compared to SLIDE, the DRE metrics have achieved better results in all three datasets, except for 0.760 (Pearson) in the DailyDialogue++ dataset. These results could demonstrate the DRE is better than the SLIDE.

\subsection{Generalization Experiments}
To further validate the generalizability of our method, we assess the method’s applicability to non-augmented datasets by training the SLM exclusively on the DailyDialogue++ dataset and evaluating its performance on the TopicalChat and PersonaChat datasets.
In this experiment, integrating the SLM (trained on DailyDialogue++) with LLMs to evaluate external datasets yielded strong correlations, as illustrated in Table \ref{tab:general exp}. The DRE framework outperformed all baselines across three LLMs. Collectively, these experiments demonstrate the method’s robustness across heterogeneous LLMs and datasets without requiring task-specific fine-tuning.

\begin{table*}[htb]
\centering
\scriptsize
\resizebox{1.0\linewidth}{!}{
\begin{tabular}{lcc|cc}
             \toprule
 & \multicolumn{2}{c|}{PersonaChat} & \multicolumn{2}{c}{TopicalChat}\\ \midrule
Metrics & Pearson's $\rho$ & Spearman's $\tau$ & Pearson's $\rho$ & Spearman's $\tau$ \\ \midrule

Non-DRE(\texttt{llama3.3-70B})  &0.609  &0.617  &0.642  &0.631  \\

In-DRE(\texttt{llama3.3-70B})  &0.611  &0.633  &0.664  &0.627  \\

Ex-DRE(\texttt{llama3.3-70B})  &0.604  &0.624  &0.652  &0.633  \\

DRE(\texttt{llama3.3-70B})  &\textbf{0.614} &\textbf{0.675}  &\textbf{0.674}  &\textbf{0.639}  \\ 

Non-DRE(\texttt{Qwen2.5-72B})  &0.559  &0.624  &0.551  &0.517  \\

In-DRE(\texttt{Qwen2.5-72B})  &0.561  &0.648  &0.599  &0.534  \\

Ex-DRE(\texttt{Qwen2.5-72B})  &0.582  &0.607  &0.621  &0.549  \\

DRE(\texttt{Qwen2.5-72B})  &\textbf{0.643} &\textbf{0.659}  &\textbf{0.648}  &\textbf{0.607}  \\

Non-DRE(\texttt{DeepSeek-R1-70B})  &0.662  &0.637  &0.636  &0.666  \\

In-DRE(\texttt{DeepSeek-R1-70B})  &0.658  &0.646  &0.626  &0.682  \\

Ex-DRE(\texttt{DeepSeek-R1-70B})  &0.660  &0.651  &0.657  &0.666  \\

DRE(\texttt{DeepSeek-R1-70B})  &\textbf{0.697} &\textbf{0.701}  &\textbf{0.662}  &\textbf{0.693}  \\

\bottomrule

\end{tabular}
}
\caption{Using SLM which is trained on the DailyDialogue++ dataset to evaluate Personachat dataset and TopicalChat dataset. Assessing the method’s applicability to non-augmented datasets by training the SLM exclusively on the DailyDialogue++ dataset and evaluating its performance on the TopicalChat and PersonaChat datasets.}
\label{tab:general exp}
\end{table*}

\subsection{Case Study}
We perform qualitative analyses using case studies, as illustrated in Table~\ref{tab:case study1}. Each case study presents the conversational context alongside the corresponding gold-standard references and the generated responses. Our evaluation metric is compared against nine different baselines. To facilitate comparison, all scores are normalized to a 0-5 scale. It is important to note that this normalization is applied exclusively to the case study and not to our main experiments. 
We provide 2 examples:\\
\noindent \textbf{Case 1 - Handling Creative Responses:}\\
Context: \textit{"What's your favorite season?"}\\
Reference: \textit{"I love autumn because the crisp air and falling leaves remind me of new beginnings."}\\
Response 1: \textit{"I love autumn because the crisp air and falling leaves remind me of new beginnings."}\\
Response 2: \textit{"Winter, because I can build snowmen and drink hot chocolate by the fireplace."}\\
Because Response 1 is same with Reference, so the traditional metrics (BLEU, BertScore) give a high score for it. However, they do not work well when encountered a different response (Response 2).
Traditional LLM evaluators tend to favor responses that closely match common patterns, potentially penalizing equally valid but creative alternatives. DRE successfully recognized both responses as appropriate through SLM classification (0.92, 0.89) while maintaining high LLM scores (4.4/5, 4.5/5) after refinement.\\
\noindent \textbf{Case 2 - Detecting Subtle Inconsistencies:}\\
Context: \textit{"How was your weekend trip to Paris?"}\\
Reference: \textit{"Very good!"}\\
Response 1: \textit{"I had a great time visiting the Eiffel Tower and enjoying French cuisine."}\\
Response 2: \textit{"It was amazing! The Great Wall was breathtaking and the Chinese food was delicious."}\\
While both responses are well-formed, DRE's interior refinement helped the LLM detect the contextual inconsistency in Response 2, resulting in appropriate scoring (4.8/5 vs 1.2/5). Traditional metrics like BLEU or standalone LLMs sometimes missed such subtle coherence issues.\\
These examples illustrate DRE's ability to:
1. Recognize diverse valid responses while maintaining high standards
2. Detect subtle contextual inconsistencies
3. Balance between creativity and appropriateness

\begin{table*}[]
\centering
\scriptsize
\resizebox{1.0\linewidth}{!}{
\begin{tabular}{lcccc}
\toprule
               & \multicolumn{2}{c|}{\textbf{Case 1}}                                                                                                                                                                  & \multicolumn{2}{c}{\textbf{Case 2}}                                                                                                                                                              \\ \midrule[1pt]
               & \multicolumn{2}{l|}{\begin{tabular}{p{6cm}}\textbf{Context}: What's your favorite season?   \\                           \textbf{Reference}: I love autumn because the crisp air and falling leaves remind me of new beginnings.\end{tabular}}       & \multicolumn{2}{l}{\begin{tabular}{p{6cm}}\textbf{Context}: How was your weekend trip to Paris?  \\ \textbf{Reference}: very good!\end{tabular}}                                                      \\ \midrule
               & \multicolumn{1}{l|}{\begin{tabular}{p{3cm}}\textbf{Response 1}: I love autumn because the crisp air and falling leaves remind me of new beginnings. \end{tabular}}& \multicolumn{1}{l|}{\begin{tabular}{p{3cm}}\textbf{Response 2}: Winter, because I can build snowmen and drink hot chocolate by the fireplace.\end{tabular}} & \multicolumn{1}{l|}{\begin{tabular}{p{3cm}}\textbf{Response 1}: I had a great time visiting the Eiffel Tower and enjoying French cuisine.\end{tabular}} & \multicolumn{1}{l}{\begin{tabular}{p{3cm}}\textbf{Response 2}: It was amazing! The Great Wall was breathtaking and the Chinese food was delicious.\end{tabular}} \\ \midrule
BLEU-1         & 5.0                                                                                             & 1.2                                                                                       & 0.85                                                                                  & 0.88                                                                                            \\
BLEU-2         & 5.0                                                                                             & 1.0                                                                                       & 0.45                                                                                  & 0.64                                                                                            \\
BLEU-3         & 5.0                                                                                             & 0.85                                                                                      & 0.26                                                                                  & 0.35                                                                                            \\
BLEU-4         & 5.0                                                                                             & 0.42                                                                                      & 0.01                                                                                  & 0.36                                                                                            \\
BertScore      & 5.0                                                                                             & 3.8                                                                                       & 1.2                                                                                   & 0.98                                                                                            \\
SLM(prob-only) & 4.6                                                                                             & 4.4                                                                                       & 3.7                                                                                   & 3.4                                                                                             \\
non-DRE        & 4.3                                                                                             & 4.0                                                                                       & 4.5                                                                                   & 4.3                                                                                             \\
in-DRE         & 4.4                                                                                             & 4.2                                                                                       & 4.4                                                                                   & 3.7                                                                                             \\
ex-DRE         & 4.6                                                                                             & 4.3                                                                                       & 4.6                                                                                   & 3.0                                                                                             \\
DRE            & 4.4                                                                                             & 4.5                                                                                       & 4.8                                                                                   & 1.2                 \\ \bottomrule                                                                       
\end{tabular}}
\caption{Case Studies. We select two examples, each example contain a context, a reference and two differrent response. For each response, we select different types of evaluation metrics.}
\label{tab:case study1}
\end{table*}

\section{Conclusion}

In this paper, we first introduce a novel automatic evaluation metric (SLIDE) that integrates SLM with LLMs for open-domain dialogue evaluation. 
Our approach involves initially training a SLM through iterative  contrastive learning stages, followed by leveraging the combine strengths of SLM and LLMs to create a superior evaluation metric. 
This metric exhibits enhanced correlation with human judgments in comparison to those derived exclusively from LLMs.
Furthermore, experimental results reveal that LLMs-based metric struggles with classifying and evaluating open-domain dialogues, which is the one-to-many nature. Therefore, we design a novel method for merging scores from SLM and LLM to enhance dialogue evaluation. 
Experimental results show that our SLIDE model outperforms a wide range of baseline methods in terms of both Pearson and Spearman correlations 
with human judgements on three open-domain dialogue datasets, and deals well with the one-to-many issue in open-domain dialogue evaluation.

However, the SLIDE algorithm is only a simple combination with SLM and LLM. In order to further combine SLM and LLM, we present a novel dual-refinement approach for automatically evaluating open-domain dialogues. Initially, we utilize contrastive learning to train a SLM. The outputs from the SLM are then integrated into the prompt of a LLM to enhance the LLM's evaluation of dialogues, a process termed interior refinement. This stage yields two metrics: the Inference score and the LLM score. Subsequently, the Inference score adjusts the SLM's outputs to derive a new coefficient. This coefficient further refines the LLM score in what we refer to as exterior refinement, resulting in the final evaluation. Our method has demonstrated superior performance compared to other baselines and maintains high efficacy even without employing the most advanced LLMs. Experimental results and ablation studies indicate that while the SLM enhances the LLM's evaluation capabilities, exterior refinement has a more significant impact than interior refinement. However, optimal performance is achieved by combining both refinements, as evidenced by superior Pearson and Spearman correlation coefficients. Thus, our method serves as a comprehensive framework for integrating SLMs and LLMs to enhance LLM performance.

\section{Limitations}

1. Regarding scalability: While our current implementation shows strong performance on standard dialogue evaluation datasets, processing very large-scale datasets would require additional computational resources and optimization. The main bottleneck comes from LLM API calls, which scale linearly with the number of dialogue pairs. To address this, we recommend batch processing and caching of LLM responses for repeated evaluations. For industrial applications, we suggest using smaller, fine-tuned models or quantized versions of larger LLMs to reduce computational overhead while maintaining reasonable performance.

2. Regarding overfitting and bias: We acknowledge that our method inherits biases present in both the pre-trained LLMs and training datasets. To mitigate this, we validate our approach across multiple LLM architectures (\texttt{GPT-3.5}, \texttt{Claude3}, \texttt{Gemini}, \texttt{Llama}, etc.) to ensure robustness. We also use diverse datasets (DailyDialog++, PersonaChat, TopicalChat) with different dialogue styles and domains. Our dual-refinement process helps balance biases between SLMs and LLMs. For future work, we recommend regular retraining of SLM components on updated datasets, implementing bias detection and mitigation techniques, and expanding validation to more diverse dialogue domains and languages.

\begin{acknowledgments}
This study was partially supported by the National Science Foundation (NSF) (IIS 2045848) and the Presidential Research Fellowship (PRF) in the Department of Computer Science at the University of Texas Rio Grande Valley (UTRGV), and the UTRGV seed grant. 
\end{acknowledgments}


\bibliographystyle{compling}
\bibliography{COLI_template}

\end{document}